\documentclass[10pt,twocolumn,letterpaper]{article}

\usepackage[pagenumbers]{cvpr} 

\usepackage{multirow}
\usepackage{float}
\usepackage{overpic}
\usepackage{caption}
\usepackage{rotating}
\usepackage{colortbl}
\usepackage{algorithm}
\usepackage{algpseudocode}
\usepackage{marvosym}
\usepackage{graphicx}
\usepackage{amsmath}
\usepackage{amssymb}
\usepackage{booktabs}

\definecolor{cvprblue}{rgb}{0.21,0.49,0.74}
\usepackage[pagebackref,breaklinks,colorlinks,allcolors=cvprblue]{hyperref}

\def\paperID{*****} 
\def\confName{CVPR}
\def\confYear{2025}

\title{MC$^2$: Multi-concept Guidance for Customized Multi-concept Generation}

\author{Jiaxiu Jiang$^1$\quad
Yabo Zhang$^1$\quad
Kailai Feng$^1$\quad
Xiaohe Wu$^1$\quad 
Wenbo Li$^2$\quad \\
Renjing Pei$^2$\quad
Fan Li$^2$\quad
Wangmeng Zuo$^1$ \textsuperscript{\Letter}\\
Harbin Institute of Technology$^1$ \qquad Huawei Noah’s Ark Lab$^2$
}

\begin{document}
\maketitle
\begin{abstract}
Customized text-to-image generation, which synthesizes images based on user-specified concepts, has made significant progress in handling individual concepts. However, when extended to multiple concepts, existing methods often struggle with properly integrating different models and avoiding the unintended blending of characteristics from distinct concepts. In this paper, we propose MC$^2$, a novel approach for multi-concept customization that enhances flexibility and fidelity through inference-time optimization. MC$^2$ enables the integration of multiple single-concept models with heterogeneous architectures. By adaptively refining attention weights between visual and textual tokens, our method ensures that image regions accurately correspond to their associated concepts while minimizing interference between concepts. Extensive experiments demonstrate that MC$^2$ outperforms training-based methods in terms of prompt-reference alignment. Furthermore, MC$^2$ can be seamlessly applied to text-to-image generation, providing robust compositional capabilities. To facilitate the evaluation of multi-concept customization, we also introduce a new benchmark, MC++. The code will be publicly available at \url{https://github.com/JIANGJiaXiu/MC-2}.
\end{abstract}    
\section{Introduction}
\label{sec:intro}

Diffusion models have facilitated advancements in creative visual generation \cite{ho2020denoising,Nichol2021GLIDETP,Ramesh2022HierarchicalTI,saharia2022photorealistic,Rombach2021HighResolutionIS,zhang2023controlvideo}.
Leveraging the power of text-to-image diffusion models, customized generation \cite{gal2022image,ruiz2023dreambooth,kumari2023multi,tewel2023key,weielite,xiao2023fastcomposer,gu2023mixofshow,po2023orthogonal,Ma2023SubjectDiffusionOD,DBLP:conf/iccv/HanLZMMY23,zhang2023compositional} enables users to craft images aligned with their personalized concepts.
While earlier endeavors \cite{gal2022image,ruiz2023dreambooth,weielite} predominantly address the intricacies of single-concept customization, the challenge persists when it comes to multi-concept customization in terms of flexibility and fidelity.

\begin{figure}[tb]
	\centering
	\includegraphics[width=\linewidth]{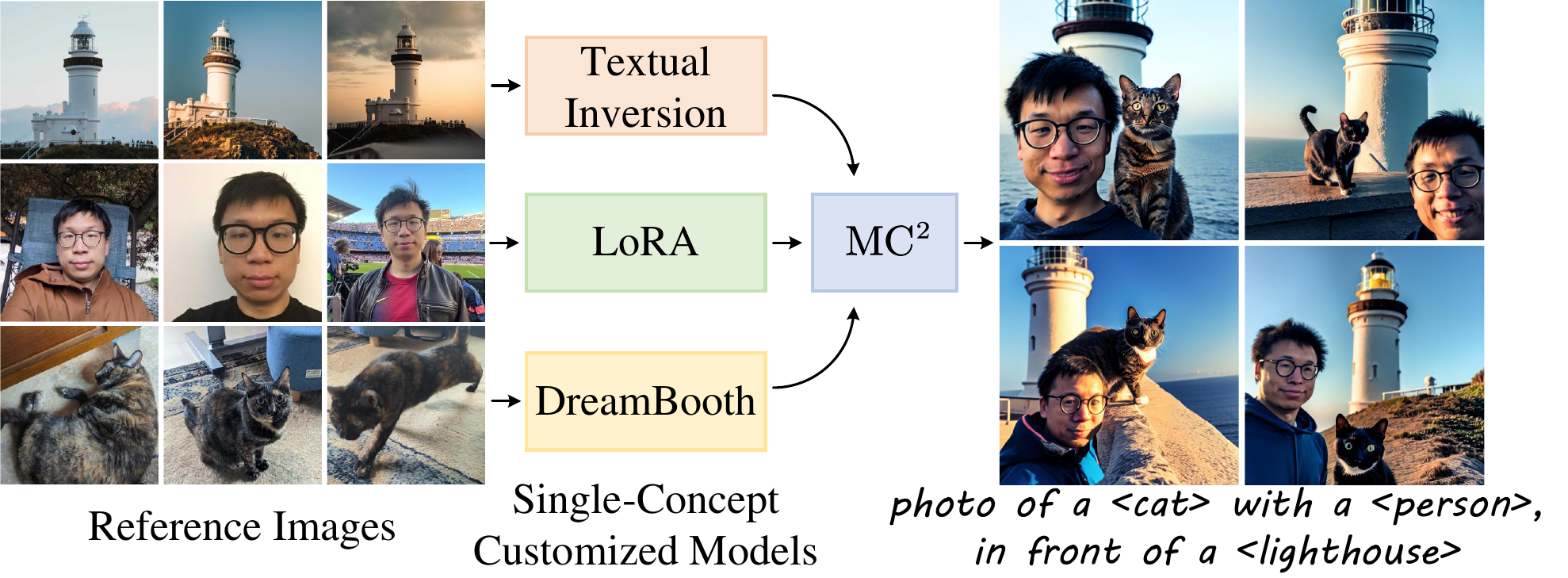}
	\caption{\textbf{Overview of our method.} 
		MC$^2$ composes separately trained customized models to generate compositions of multiple customized concepts. Here, we train a Textual Inversion \cite{gal2022image} model for \texttt{<lighthouse>}, a LoRA \cite{hu2022lora} for \texttt{<person>}, and a DreamBooth \cite{ruiz2023dreambooth} model for \texttt{<cat>} with the reference images of each concept. The reference images are from the CustomConcept101 dataset\cite{kumari2023multi}.
	}
	\label{fig:teaser}
\end{figure}

Existing works performs customized multi-concept generation through joint training \cite{kumari2023multi,DBLP:conf/iccv/HanLZMMY23} or merging single-concept customized models \cite{tewel2023key,po2023orthogonal,DBLP:conf/icml/LiuFZZZLZZC23,gu2023mixofshow}.
Joint training \cite{kumari2023multi, DBLP:conf/iccv/HanLZMMY23} requires access to the reference images of the specified concepts, and retrain all parameters when introducing a new concept.
Merging single-concept customized models \cite{tewel2023key,po2023orthogonal,DBLP:conf/icml/LiuFZZZLZZC23,gu2023mixofshow} promisingly reduces the training cost.
Nonetheless, it is challenging for the above methods to integrate multiple heterogeneous single-concept customized models, as they require specific network architectures and access to training data.
Users may obtain a large amount of single-concept customized models from the community, but cannot acquire their training data in most cases.
Therefore, these models are difficult to be integrated to generate compositions of multiple customized concepts.
Albeit Cones~2~\cite{liu2023cones} does not require joint training or model merging for generating composition of customized concepts, it introduces additional and inconvenient condition signals, \eg, bounding boxes or segmentation masks.
Encoder-based customization methods \cite{Ma2023SubjectDiffusionOD,xiao2023fastcomposer} usually involve finetuning the pretrained text-to-image diffusion model to accept identity information extracted from the reference images, which are orthogonal to our work.

To address the above mentioned problems, we propose the \textbf{M}ulti-\textbf{C}oncept guidance for customized \textbf{M}ulti-\textbf{C}oncept generation, termed \textbf{MC$\mathbf{^2}$}. Our proposed MC$^2$ facilitates the seamless integration of separately trained heterogeneous single-concept customized models, allowing for the natural synthesis of a composition of their distinct concepts without additional training. As shown in \cref{fig:teaser}, a Textual Inversion \cite{gal2022image} model, a LoRA \cite{hu2022lora} and a DreamBooth \cite{ruiz2023dreambooth} model are combined with our method, demonstrating its versatility without requiring extra training or additional conditioning information. Multi-concept guidance (MCG) acts as an avenue for the separately trained customized models to communicate with each other. In each step of MC$^2$, each customized model denoises the same latent noise map respectively, then MCG attempts to identify the activated regions of each concept and spatially disentangle them. When the activated regions of each concept have little overlap, the multiple customized concepts are more likely to be generated simultaneously with less incorrect attribute binding. Inspired by \cite{DBLP:conf/acl/TangLPJYKSLT23, Chefer2023AttendandExciteAS, ni2023ref}, we extract the cross-attention maps from the diffusion process. Certain cross-attention maps indicate the activated regions of each customized concept.
Then MCG adaptively refines the attention weights between visual and textual tokens, directing image regions to focus on their associated words while diminishing the impact of irrelevant ones.
Moreover, MC$^2$ can be extended to enhance the compositional generation ability of existing text-to-image diffusion models, yielding appealing results.

Our contributions are as follows:
\begin{itemize}
	\item We propose a novel method MC$^2$ for integrating various single-concept customized models to synthesize composition of different customized concepts, which does not require joint training, model merging or additional condition information.
	\item We construct a benchmark MC++ for evaluating multi-concept customization. The benchmark contains compositions of two to four concepts.
	\item With minor adjustments, our method can be extended to enhance the compositional generation ability of existing text-to-image diffusion models.
	\item Extensive qualitative and quantitative evaluations demonstrate that our proposed method significantly improves the performance of customized multi-concept generation and compositional text-to-image generation, even surpassing previous methods that require additional training.
\end{itemize}

\section{Related Work}
\label{sec:relatedwork}

\textbf{Customized multi-concept generation.}
Customized generation methods \cite{gal2022image, ruiz2023dreambooth, weielite} aim to generate images of a specified concept given a few user-provided images of the concept. Custom Diffusion \cite{kumari2023multi} first proposes to extend the task to a multi-concept scenario, where the compositions of multiple customized concepts are expected. Custom Diffusion finetunes a diffusion model given reference images of multiple concepts or merges single-concept customized diffusion models with constrained optimization. SVDiff \cite{DBLP:conf/iccv/HanLZMMY23} proposes the data augmentation strategy Cut-Mix-Unmix which splices two reference images together for finetuning the diffusion model. Some works \cite{tewel2023key, po2023orthogonal, DBLP:conf/icml/LiuFZZZLZZC23} separately train customized models for each concept then merge the models to get one customized model with their proposed formulas. Cones \cite{DBLP:conf/icml/LiuFZZZLZZC23} additionally requires finetuning for the merged customized model for better generation quality. Mix-of-Show \cite{gu2023mixofshow} trains ED-LoRAs for each concept then merges them into one ED-LoRA via gradient fusion. Cones 2~\cite{liu2023cones} finetunes the text encoder for each concept, then derives a token embedding for it. 
Cones 2 only uses the embeddings for each concept at inference. 
Compositional Inversion \cite{zhang2023compositional} finetunes token embeddings for each concept then composes them via their proposed spatial inversion. 
Encoder-based customization methods FastComposer \cite{xiao2023fastcomposer} and Subject-Diffusion \cite{Ma2023SubjectDiffusionOD} also support multi-concept customization, which are orthogonal to our work.
VideoDreamer \cite{chen2023videodreamer} and CustomVideo \cite{wang2024customvideo} consider multi-concept customization for text-to-video generation with diffusion models. DreamMatcher \cite{nam2024dreammatcher} enhances existing customization methods with semantic matching. Concurrent work \cite{zhong2024multi} proposes to compose multiple LoRAs for image generation, but not particularly for multi-concept customization.
Another line of work \cite{avrahami2023break, jin2023image, agarwal2023image} focuses on extracting multiple concepts or attributes from a single image. Break-A-Scene~\cite{avrahami2023break} proposes a masked diffusion loss and a cross-attention loss to encourage the disentanglement between the concepts during training.

MC$^2$ supports combining multiple separately trained customized models of different architectures, \eg Textual Inversion~\cite{gal2022image}, LoRA~\cite{hu2022lora} and DreamBooth~\cite{ruiz2023dreambooth}, without additional training or layout information.

\noindent
\textbf{Compositional generation.}
Text-to-image diffusion models sometimes fail to generate all the subjects mentioned in the input text prompt, and the involved attributes may be bound to incorrect subjects. Previous works \cite{liu2022compositional, BarTal2023MultiDiffusionFD, feng2022training, Chefer2023AttendandExciteAS, DBLP:conf/iccv/AgarwalKJSGS23, rassin2023linguistic, meral2023conform, kang2023semantic, zhang2024realcompo} aim to address the problem without additional training. \cite{liu2022compositional, BarTal2023MultiDiffusionFD} adopt the architecture of parallel diffusion models. \cite{feng2022training} strengthens the semantic information of some selected word of the prompt. \cite{Chefer2023AttendandExciteAS, DBLP:conf/iccv/AgarwalKJSGS23, rassin2023linguistic, meral2023conform} optimize the noise map to encourage the co-occurrence of the mentioned concepts. \cite{kang2023semantic} modulates the guidance direction of diffusion models during inference. Concurrent work RealCompo~\cite{zhang2024realcompo} enhances text-to-image diffusion models with layout-to-image diffusion models to promote compositionality of the generated images, which is orthogonal to our work.

\noindent
\textbf{Sampling guidance for diffusion models.}
Guidance methods interfere with the sampling process of the diffusion models to achieve various effects. Some works \cite{dhariwal2021diffusion, ho2021classifier, Nichol2021GLIDETP, hong2023improving} improve the sampling quality of pretrained diffusion models or make the generation results more consistent with the input condition. Others \cite{bansal2023universal, kim2023dense, ge2023expressive, chen2024training, epstein2024diffusion} enable additional condition input for pretrained text-to-image diffusion models, to control layout, style or other attributes of the generated images.

\begin{figure*}[tb]
	\centering
	\includegraphics[width=0.95\linewidth]{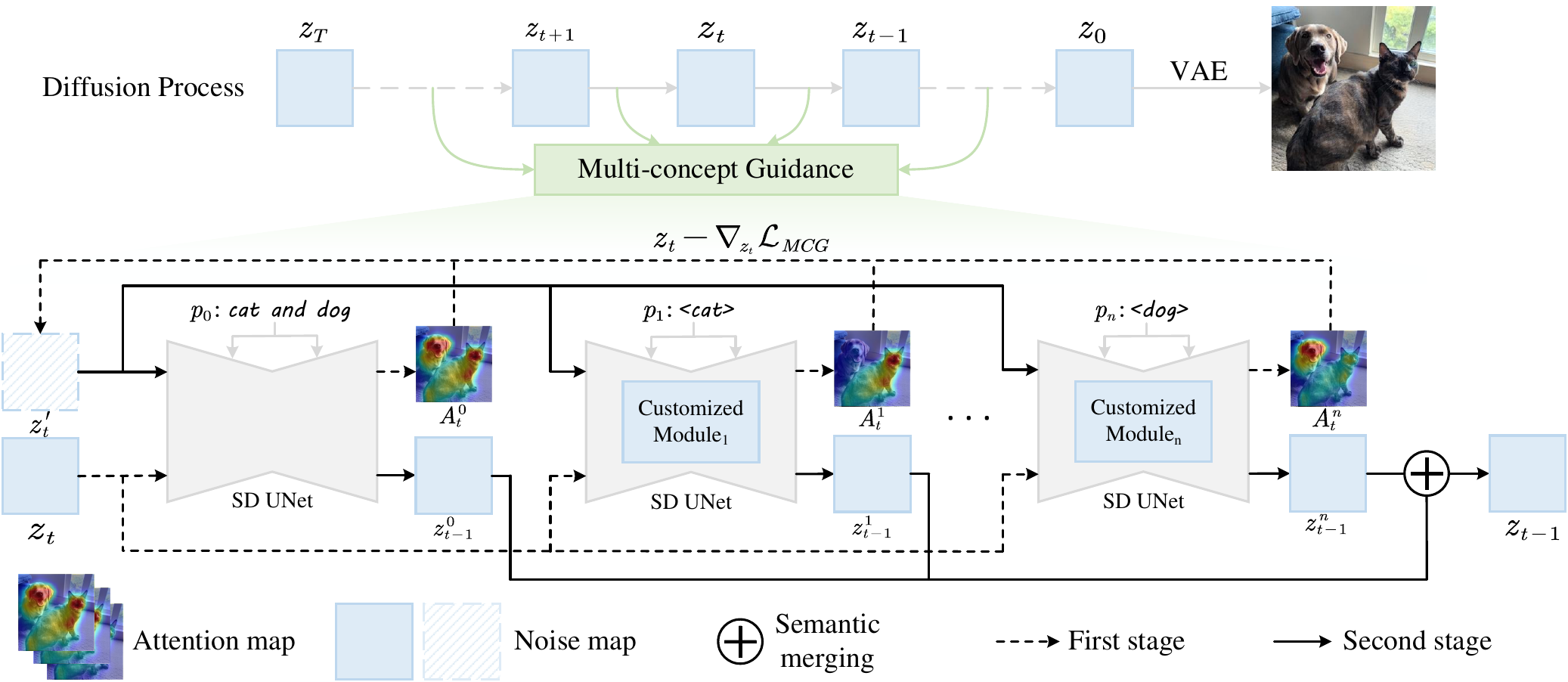}
	\caption{\textbf{Illustration of our proposed MC$\mathbf{^2}$.} Multi-concept Guidance (MCG) is performed at each step of the diffusion process. In the first stage, several parallel diffusion models with different customized modules take the same noise map $z_t$ as input. $p_0$, $p_1$ and $p_n$ denote text prompts encoded by the CLIP text encoder. Then the cross-attention maps of certain tokens are extracted to compute the $\mathcal{L}_{MCG}$ to update $z_t$. In the second stage, the diffusion models take $z_t'$ as input and $z_{t-1}$ is calculated via semantic merging. When omitting the customized modules and substituting $\mathcal{L}_{CompGen}$ for $\mathcal{L}_{MCG}$, the framework applies to plain compositional generation.
	}
	\label{fig:overview}
\end{figure*}

\section{Method}

\subsection{Problem Definition}
\label{sec:definition}

In the realm of customized multi-concept generation, the goal is to create images encompassing multiple user-specified concepts. Specifically, we focus on a scenario where we have access solely to the trained single-concept customized models $\{\mathcal{G}_1, \mathcal{G}_2, \dots, \mathcal{G}_n\}$ for each concept. In this context, the reference images for each concept are not available. Our goal is to propose some structure $\mathcal{S}$, which composes the single-concept customized models to enable the synthesis of an image $x$ that incorporates the $n$ concepts. Namely we aim to construct $\mathcal{S}$, such that:
\begin{equation}
	x = \mathcal{S}(\mathcal{G}_1, \mathcal{G}_2, \dots, \mathcal{G}_n)(p),
	\label{eq:nG}
\end{equation}
where $p$ is the input text prompt.

\subsection{Preliminaries}

Recent text-to-image diffusion models are typically based on Denoising Diffusion Probabilistic Models (DDPMs). DDPMs model image generation as a denoising process where an individual denoising step is formulated as:
\begin{equation}
	x_{t-1} = x_{t} - \epsilon_{\theta}(x_{t}, t) + \mathcal{N}(0, \sigma_{t}^{2}I),
\end{equation}
where image $x_{t-1}$ is a denoised version of $x_{t}$, $\epsilon_{\theta}$ is a model to predict the noise in $x_{t}$, $\mathcal{N}$ is a Gaussian distribution with learned covariance matrix $\sigma_{t}^{2}I$. The denoising step is performed $T$ times to get the final output image $x_0$. $x_T$ is sampled from a Gaussian prior. $\epsilon_{\theta}(x_{t}, t)$ is also called the score function of the diffusion model.

Composable Diffusion~\cite{liu2022compositional} shows that diffusion models are functionally similar to Energy Based Models (EBMs). Multiple EBMs can be composed to get a new EBM that models a composed distribution of the distributions modeled by the multiple EBMs. A trained diffusion model $\epsilon_{\theta}(x_{t}, t)$ can be viewed as an implicitly parameterized EBM. Then multiple diffusion models can be composed to obatin a new diffusion model. Let model $\mathcal{G}_i$ contain concept $c_i$, then an image containing $c_i$ can be sampled using the score function $\epsilon_{\theta}(x_{t}, t|c_i)$ of the conditional distribution $p(x|c_i)$. We then sample from the conditional distribution $p(x|c_1,\dots, c_n)$ with the composed score function $\hat{\epsilon}_{\theta}(x_{t}, t)$:
\begin{equation}
	\hat{\epsilon}_{\theta}(x_{t}, t) = \epsilon_{\theta}(x_{t}, t) + \displaystyle\sum_{i=1}^{n}w_i(\epsilon_{\theta}^{i}(x_{t}, t|c_i) - \epsilon_{\theta}(x_{t}, t)),
	\label{eq:ComposedDiffusion}
\end{equation}
where $w_i$ is a hyperparameter corresponding to the temperature scaling on concept $c_i$.
For more detailed derivation, please refer to \cite{liu2022compositional}.

\begin{figure}[tb]
	\centering
	\includegraphics[width=\linewidth]{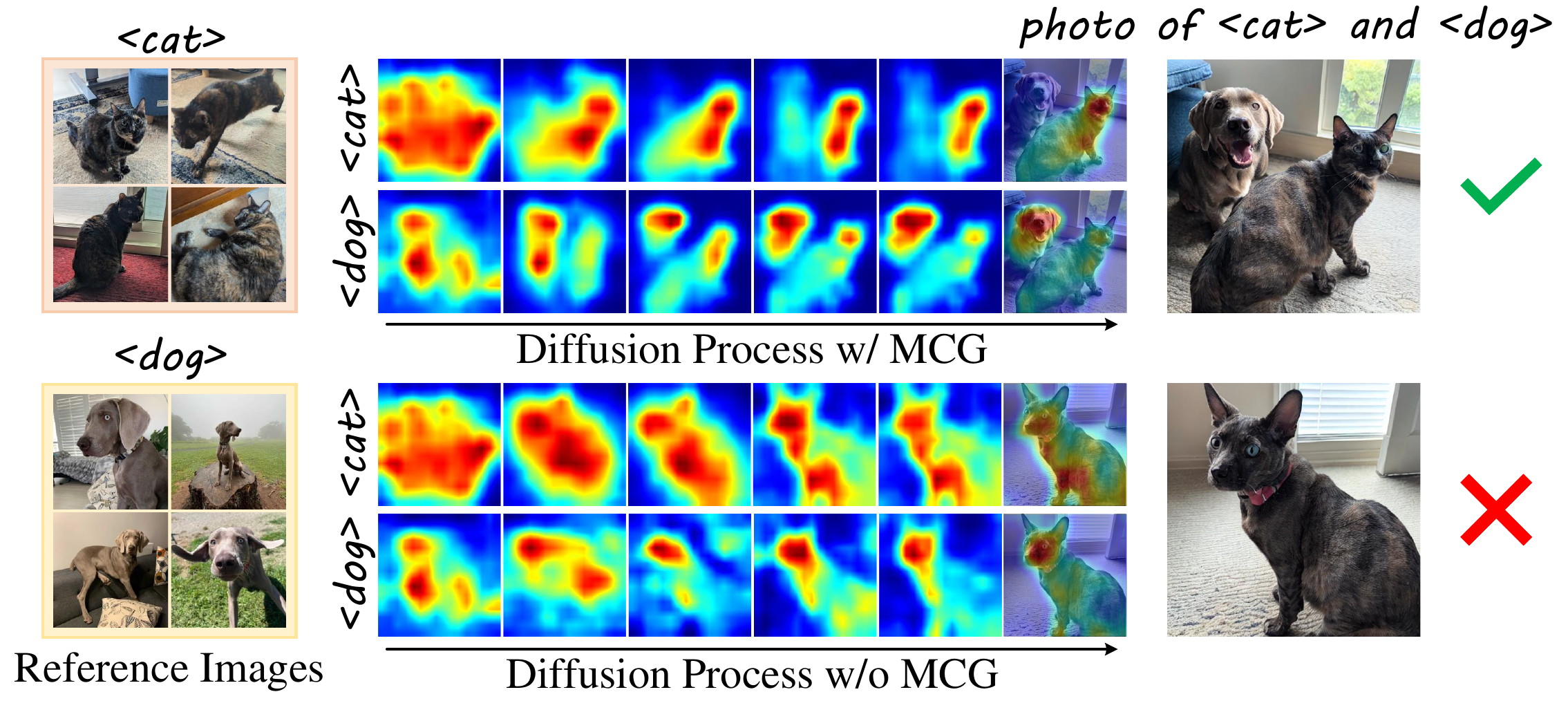}
	\caption{\textbf{Visualization of MCG.}
		MCG adaptively refines the attention weights between visual and textual tokens, directing image regions to focus on their associated words while diminishing the impact of irrelevant ones.
	}
	\label{fig:attn}
\end{figure}

\Cref{eq:ComposedDiffusion} can act as the structure $\mathcal{S}$ in \cref{eq:nG}, but it may confuse objects' attributes or create an object that merges the specified concepts, as shown in the second row of \cref{fig:attn} and the first row of \cref{fig:comparison-gsn}. Composable LoRA~\cite{composablelora} proposes to use the structure with LoRAs \cite{lorafordiffusion}, \ie to merge the concepts contained in different LoRA modules. As shown in \cref{fig:overview}, we adopt the architecture of multiple parallel diffusion models with different customized modules to start with. Limited by \cite{liu2022compositional}, the diffusion models are the instances of the same model. With this architecture, the customized modules can be of different forms, \eg Textual Inversion \cite{gal2022image}, LoRA \cite{hu2022lora} and DreamBooth \cite{ruiz2023dreambooth}, as they are not required to be merged to one model, shown in \cref{fig:teaser}.

\subsection{Multi-concept Guidance}
Our proposed MC$^2$ performs Multi-concept Guidance (MCG) at each step of the diffusion process.
At each step of the diffusion process we refine the attention weights between visual and textual tokens in order to spatially disentangle the customized concepts.
We define a loss objective to measure the extent to which the customized concepts are spatially disentangled. As illustrated in \cref{fig:overview}, the noised latent map $\mathrm{z}_t$ is updated with the objective in the first stage. In the second stage, the diffusion models take the updated noised latent map $z_t'$ as input and $z_{t-1}$ is produced via semantic merging for the next denoising step.

We attempt to measure the extent to which the customized concepts are spatially disentangled with the help of the cross-attention maps generated at each step of the diffusion process. They can be regarded as features showing how strongly each latent image patch is associated with each text token embedding. Each cross-attention layer in the unet takes latent image features $f \in \mathbb{R}^{(h\times w)\times l}$ and text features $c \in \mathbb{R}^{s\times d}$ as input. After linear layer $Q = W^Q f$, $K= W^K c$, $V= W^V c$, the attention maps are then:
\begin{equation}
	A=\mathrm{Softmax}(\frac{Q K^T}{\sqrt{d'}})\in \mathbb{R}^{(h\times w)\times s}
	\label{eq:crossattention}
\end{equation}

As Shown in \cref{fig:overview}, we first construct $n + 1$ sub-prompts $\{p_0, p_1, \dots, p_n\}$ for the diffusion models, where $n$ denotes the number of customized concepts. Each concept is customized by a customized module. $p_0$ describes the image to be generated as a whole. $p_1$ to $p_n$ describe each customized concept. The \emph{trigger words} of the customized modules are included in their sub-prompts.
Single-concept models typically have their trigger words for activating the customized information during the generation process.
For Textual Inversion \cite{gal2022image}, the trigger word is the trained token embedding itself.
For DreamBooth \cite{ruiz2023dreambooth}, the trigger words are some rare tokens selected from the vocabulary.
We extract the cross-attention maps corresponding to the trigger words as they indicate whether and where the customized concepts are activated in the generation. Different cross-attention layers in the Stable Diffusion produce attention maps of different resolutions.
Following \cite{Chefer2023AttendandExciteAS}, we empirically choose the $16 \times 16$ attention maps. Then we average all the attention layers and heads to get the final attention map $A\in \mathbb{R}^{h\times w}$.

When a trigger word of a customized concept contains multiple tokens, each token may correspond to different part of the concept. Intuitively, we want the activated regions to stay close to each other, or to have more overlap. Inspired by \cite{DBLP:conf/iccv/AgarwalKJSGS23}, we encourage the IoU of the activated regions of the trigger tokens to be higher. Let $A_i^k \in \mathbb{R}^{h\times w}$ represent the attention map corresponding to the $i$-th token within the $k$-th sub-prompt, while $S_k$ signifies the collection of trigger token indices associated the $k$-th sub-prompt. Our proposed \emph{intra-prompt aggregation loss} is defined as:
\begin{equation}
	\mathcal{L}_{intra}=\frac{1}{n}\displaystyle\sum_{k=1}^{n}\frac{2}{(\left| S_k\right| - 1)\left| S_k\right|}\displaystyle\sum_{\substack{i, j\in S_k \\\forall i<j}}(1-\frac{\min(A_i^k,A_j^k)}{A_i^k+A_j^k}),
	\label{eq:intraprompt}
\end{equation}
where the $\min$ operation is performed at the pixel dimension.

To spatially disentangle different customized concepts, we define another loss term to reduce overlap of the regions activated by different customized modules. For sub-prompt $p_k$, we first aggregate the attention maps of the trigger tokens to get one attention map. Here, we empirically average the attention maps. After being smoothed by a Gaussian filter, we get the final attention map $A^k \in \mathbb{R}^{h\times w}$ indicating the activated region of the customized concept. \Cref{fig:attn} visualizes the attention maps of the involved concepts. Our proposed \emph{inter-prompt disentanglement loss} is then:
\begin{equation}
	\mathcal{L}_{inter}=\frac{2}{(n-1)n}\displaystyle\sum_{0<i<j\leqslant n}\frac{\min(A^i, A^j)}{A^i + A^j}.
	\label{eq:interprompt}
\end{equation}
The overall loss objective is a linear combination of $\mathcal{L}_{inter}$ and $\mathcal{L}_{intra}$ with coefficient $\alpha$, and $z_t$ is updated to $z_t'$ with learning rate $\lambda$:
\begin{gather}
	\mathcal{L}_{MCG}=\mathcal{L}_{inter}+\alpha \cdot \mathcal{L}_{intra},\label{eq:loss} \\
	z_t' = z_t - \lambda \cdot \nabla_{z_t} \mathcal{L}_{MCG}. 
\end{gather}
Then the diffusion models take $z_t'$ as input, and $z_{t-1}$ is calculated via semantic merging:
\begin{equation}
	z_{t-1} = z_{t-1}^{u} + \displaystyle\sum_{i=0}^{n}w_i(z_{t-1}^{i}-z_{t-1}^{u}).
	\label{eq:merging}
\end{equation}
The coefficients $\{w_i\}_n$ controls the intensity of the customized concepts. \Cref{eq:merging} can be seen as a implementation of \cref{eq:ComposedDiffusion}, or be regarded as an extension of classifier-free guidance\cite{ho2021classifier}. The unconditional output $z_{t-1}^u$ is produced by the uncustomized diffusion model with blank text input, which is omitted in \cref{fig:overview} for simplicity.

Our method can be extended to enhance the compositional capabilities of text-to-image generation. The customized modules are omitted as customization is not needed in such setting. Besides $\mathcal{L}_{intra}$ and $\mathcal{L}_{inter}$, we also include $\mathcal{L}_{A\&E}$ proposed by \cite{Chefer2023AttendandExciteAS}. $\mathcal{L}_{A\&E}$ attempts to maximize the attention values for each subject token so as to encourage the occurrence of the subject. For a subject name that has multiple tokens, we empirically average the attention maps corresponding to each of them. Then the loss objective for compositional generation is defined as:
\begin{gather}
	\mathcal{L}_{CompGen} = \mathcal{L}_{A\&E}+\alpha_1 \cdot \mathcal{L}_{intra} + \alpha_2 \cdot \mathcal{L}_{inter}, \label{eq:lossgsn}\\
	\mathcal{L}_{A\&E} = \max_{A\in S_A}{(1 - \max{(A)})},
\end{gather}
where $S_A$ is the set of attention maps corresponding to the subject names.

\subsection{Attention Grounding Training}

\begin{figure}
	\centering
	\includegraphics[width=\linewidth]{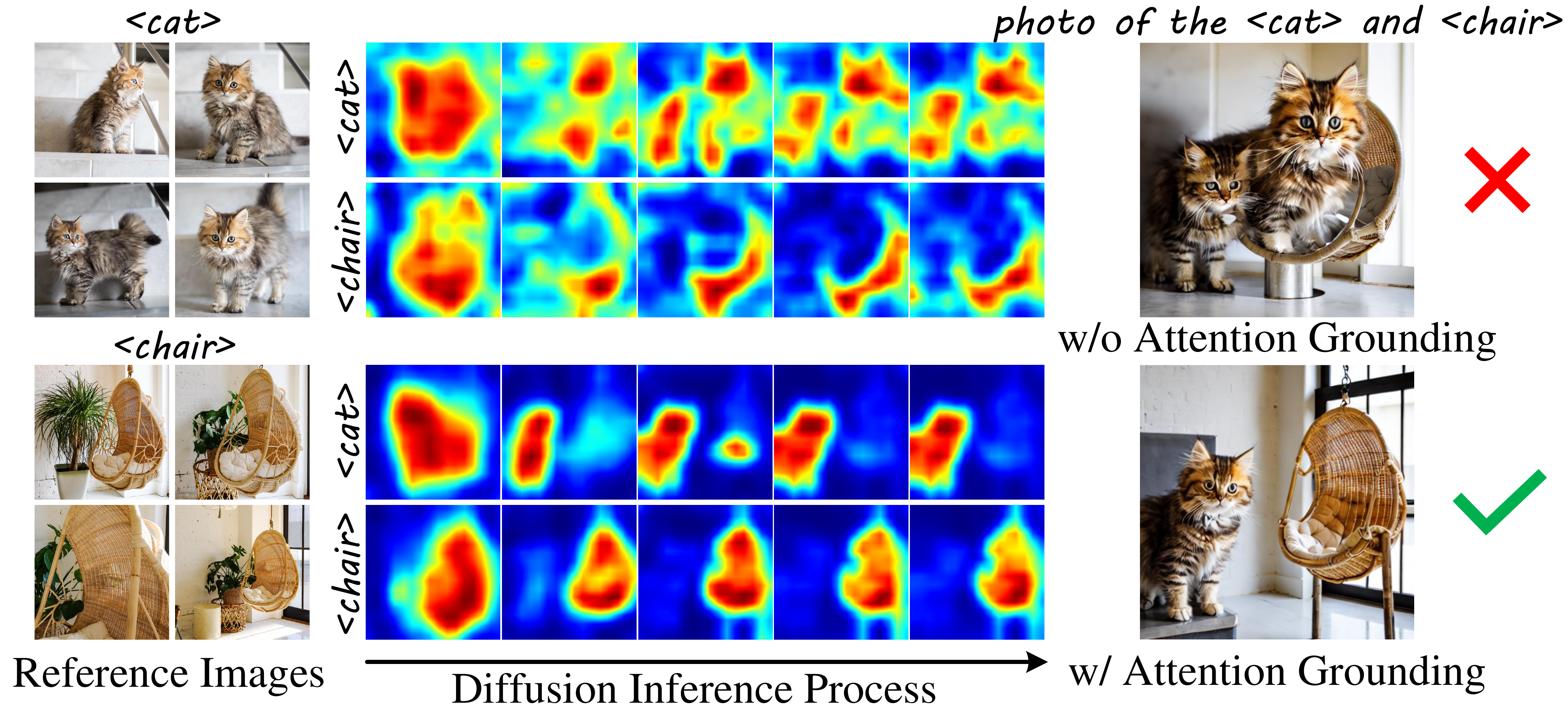}
	\caption{\textbf{Effect of attention grounding.} With attention grounding training, the concepts have more clear attention maps.
	}
	\label{fig:grounding}
\end{figure}

Not all concepts have a clear correspondence between the trigger token and the activated area. As shown in the first row of \cref{fig:grounding}, when the corresponding attention maps of \texttt{<cat>} and \texttt{<chair>} are somewhat messy, the MCG will not take the optimal effect. We incorporate two loss terms to explicitly strengthen the correspondence between the trigger tokens and the subject. The two loss terms are proposed by \cite{wang2024tokencompose} to finetune the diffusion model for better performance on generating multiple subjects. We adapt them for  single-concept customization learning. 
\begin{gather}
	\label{eq:grounding1}
	\mathcal{L}_1 = \frac{1}{s}\displaystyle\sum_{i=0}^{s}(1-\frac{\text{sum}(M_i\odot A_i)}{\text{sum}(A_i)}), \\
	\mathcal{L}_2 = -\frac{1}{shw}\displaystyle\sum_{i=0}^{s}\text{sum}(M_i\odot\log A_i + (1-M_i)\odot\log(1-A_i)).
	\label{eq:grounding2}
\end{gather}

In \cref{eq:grounding1}, $M_i$ represents a binary mask for the cross-attention map $A_i$ of the $i$-th token. \Cref{eq:grounding1} and \cref{eq:grounding2} both encourage the model to have higher attention score within the mask region of the subjects. To enable the two loss terms, we produce masks for the reference images with SAM~\cite{kirillov2023segment, ren2024grounded}. The single-concept models are trained with a linear combination of the diffusion loss term, $\mathcal{L}_1$ and $\mathcal{L}_2$. As shown in the second row of \cref{fig:grounding}, the concepts have more clear attention maps.
\section{Experiments}

\subsection{MC++ Benchmark}
\label{benchmark}

We construct a new benchmark for multi-concept customization, termed MC++. The previous benchmark CustomConcept101~\cite{kumari2023multi} contains of 101 concepts with 3-15 images for each concept. For evaluating multi-concept customization, the CustomConcept101 contains prompts for 101 compositions of two subjects. Each composition has 12 corresponding prompts, \eg \texttt{photo of the \{0\} and \{1\}}. However, the CustomConcept101 is not sufficient for evaluating compositions of more than two concepts. To evaluate the performance of customization methods on composition of more than two subjects, we collect prompts for compositions of three and four concepts. The MC++ contains 2,064 prompts in total. For details of the MC++, please refer to \textit{suppl}.

\subsection{Experiment Setup}
\label{sec:setup}

\textbf{Datasets.}
\label{sec:dataset}
For customized multi-concept generation, we perform experiments on the MC++ benchmark.

For compositional generation, we perform experiments on the benchmark proposed by \cite{Chefer2023AttendandExciteAS}. It contains three types of text prompts: (i) \texttt{a <animal A> and a <animal B>}, (ii) \texttt{a <animal> and a <color><object>}, and (iii) \texttt{a <color A><object A> and a <color B><object B>}. The dataset includes 276 prompts in total.

\begin{figure*}
	\centering
	\begin{overpic}[width=\linewidth]{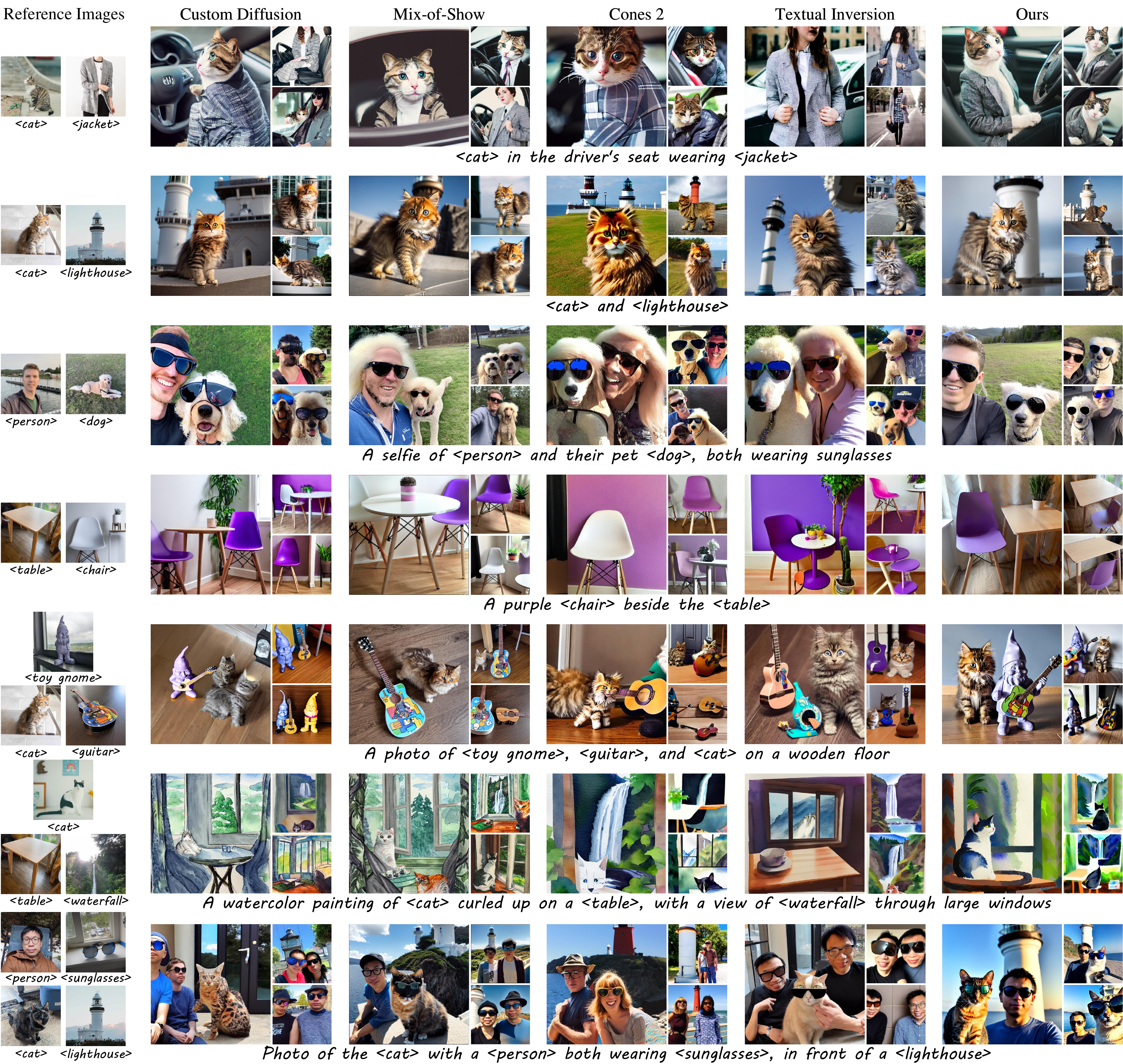}
		\scriptsize
		\put(27, 93){\cite{kumari2023multi}}
		\put(43.2, 93){\cite{gu2023mixofshow}}
		\put(59.2, 93){\cite{liu2023cones}}
		\put(79.7, 93){\cite{gal2022image}}
	\end{overpic}
	\caption{\textbf{Qualitative comparisons of customized multi-concept generation methods.} In the leftmost column are concept reference images. Only one image is shown here for each concept. The models are trained with more images. Our method demonstrates more consistency with the reference images compared to the competing methods. The competing methods sometimes omit the specified concepts.
	}
	\label{fig:comparison}
\end{figure*}

\begin{figure}
	\centering
	\begin{overpic}[width=\linewidth]{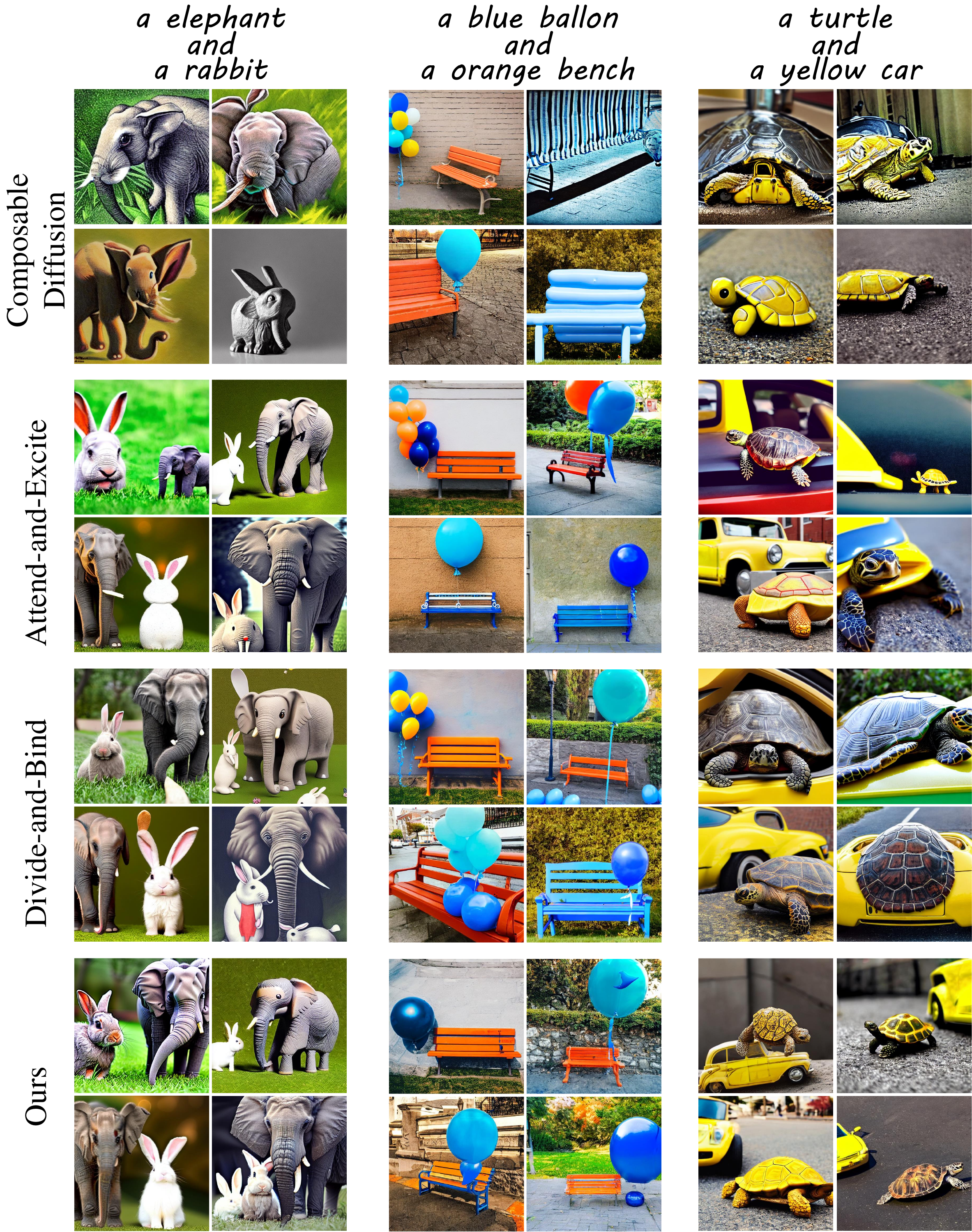}
		\scriptsize
		\begin{sideways}
			\put(42, -3.5){\cite{DBLP:conf/bmvc/LiK0K23}}
			\put(66.5, -3.5){\cite{Chefer2023AttendandExciteAS}}
			\put(86.5, -3.5){\cite{liu2022compositional}}
		\end{sideways}
	\end{overpic}
	\caption{\textbf{Qualitative comparisons of compositional generation techniques.} For each prompt, we show four images generated by each of the four considered methods where we use the same set of seeds across all approaches.
	}
	\label{fig:comparison-gsn}
\end{figure}

\noindent
\textbf{Evaluation metrics.}
For customized multi-concept generation, subject fidelity and prompt fidelity are evaluated, following \cite{ruiz2023dreambooth, kumari2023multi}. For subject fidelity, we compute two metrics: CLIP-I \cite{gal2022image} and DINO \cite{ruiz2023dreambooth}. CLIP-I is the average cosine similarity between CLIP \cite{radford2021learning} embeddings of generated and real images. The DINO metric is the average pairwise cosine similarity between the ViT-S/16 DINO \cite{caron2021emerging} embeddings of generated and real images. For a composition of two concepts, CLIP-I and DINO metrics are computed for each concept respectively then averaged. Prompt fidelity is measured as the average cosine similarity between prompt and image CLIP embedding, denoted as CLIP-T\cite{ruiz2023dreambooth}. The placeholder token in the text prompt template is replaced with the proper concept category.

For compositional generation, we evaluate full prompt similarity, minimum object similarity and text-text similarity, following \cite{Chefer2023AttendandExciteAS}. Full Prompt Similarity is similar to CLIP-T. To compute the Minimum Object Similarity, each prompt is split into two sub-prompts, each containing a single concept. We then compute the CLIP similarity between each sub-prompt and each generated image. Then the smaller of the two scores across all prompts are averaged to get the Minimum Object Similarity. To compute the text-text similarity, we employ BLIP \cite{li2022blip} to generate captions for synthesized images, then measure the CLIP similarity between the original prompt and all captions.

\noindent
\textbf{Baselines.}
For customized multi-concept generation, we compare our method with Custom Diffusion~\cite{kumari2023multi}, Mix-of-Show~\cite{gu2023mixofshow},  Cones 2~\cite{liu2023cones} and Textual Inversion~\cite{gal2022image}.
Custom Diffusion optimizes the cross-attention key and value parameter in the unet plus a token embedding given reference images.
Mix-of-Show trains ED-LoRA for each concept, then fuses multiple ED-LoRAs into one via gradient fusion. 
Mix-of-Show introduces regionally controllable sampling to control the location of each concept, which requires user-input bounding boxes for each concept, thus is omitted in our experiment.
Cones 2 finetunes the text encoder for each concept. Then a residual token embedding is derived from the text encoder for inference.
It proposes a layout guidance which requires user-input location information. The layout guidance is omitted in our experiment.

For compositional generation, we compare our method with Attend-and-Excite \cite{Chefer2023AttendandExciteAS}, Divide-and-Bind \cite{DBLP:conf/bmvc/LiK0K23} and Composable Diffusion \cite{liu2022compositional}.
Attend-and-Excite proposes a loss objective to maximize the attention values for each subject token.
Divide-and-Bind proposes a attendance loss and a binding loss to better incorporate the semantics.

\noindent
\textbf{Implementation details.}
\label{sec:implementation}
We use the Stable Diffusion v1-5 in our experiments. Without loss of generality, we choose LoRA as the single-concept customized model, as different single-concept customization methods basically finetune some selected part of the diffusion model with similar loss functions. 
We use DPM++ sampler \cite{lu2022dpm} with 30 sampling steps. Given a prompt template, \eg \texttt{photo of the \{0\} and \{1\}}, we construct the sub-prompts for a composition as follows: (i) \texttt{photo of the <cate1> and <cate2>}, (ii) \texttt{photo of the <concept1> and <cate2>, a <concept1>}, (iii) \texttt{photo of the <cate1> and <concept2>, a <concept2>}. \texttt{<cate1>} is the category name of \texttt{<concept1>} defined by CustomConcept101, \eg \texttt{dog} for \texttt{pet\_dog1}. The detailed hyperparameter setting is in \textit{suppl}.

\begin{table*}
	\scriptsize
	\centering
	\begin{tabular}{cccc|ccc|cccccc}
		\toprule
		\multirow{2}{*}{\textbf{Method}} & \multicolumn{3}{c|}{\textbf{Four concepts}} & \multicolumn{3}{c|}{\textbf{Three concepts}} & \multicolumn{3}{c}{\textbf{Two concepts}} & \multicolumn{3}{||c}{\textbf{Overall}} \\
		~ & CLIP-T\textuparrow & \makebox[0.04\textwidth]{CLIP-I\textuparrow} & DINO\textuparrow & CLIP-T\textuparrow & \makebox[0.04\textwidth]{CLIP-I\textuparrow} & DINO\textuparrow & CLIP-T\textuparrow & \makebox[0.04\textwidth]{CLIP-I\textuparrow} & DINO\textuparrow & \multicolumn{1}{||c}{CLIP-T\textuparrow} & \makebox[0.04\textwidth]{CLIP-I\textuparrow} & DINO\textuparrow \\
		\midrule
		Custom Diffusion \cite{kumari2023multi}  & 0.753 & 0.602 & 0.236 & 0.800 & 0.606 & 0.269 & \underline{0.789} & 0.645 & 0.339 & \multicolumn{1}{||c}{0.781} & 0.618 & 0.281 \\
		Mix-of-Show \cite{gu2023mixofshow}       & 0.724 & \underline{0.635} & \textbf{0.288} & 0.740 & \underline{0.635} & \underline{0.320} & 0.728 & \underline{0.670} & \underline{0.394} & \multicolumn{1}{||c}{0.731} & \underline{0.647} & \underline{0.334} \\
		Cones 2 \cite{liu2023cones}              & \underline{0.807} & 0.618 & 0.244 & \textbf{0.806} & 0.602 & 0.244 & \textbf{0.792} & 0.629 & 0.295 & \multicolumn{1}{||c}{\underline{0.802}} & 0.616 & 0.258 \\
		Textual Inversion \cite{gal2022image}    & 0.656 & 0.618 & 0.244 & 0.753 & 0.621 & 0.280 & 0.746 & 0.648 & 0.322 & \multicolumn{1}{||c}{0.718} & 0.629 & 0.282 \\
		\rowcolor{gray!15}
		Ours  & \textbf{0.825} & \textbf{0.638} & \underline{0.268} & \underline{0.805} & \textbf{0.645} & \textbf{0.324} & 0.780 & \textbf{0.692} & \textbf{0.416} & \multicolumn{1}{||c}{\textbf{0.803}} & \textbf{0.658} & \textbf{0.336} \\
		\bottomrule
	\end{tabular}
	\caption{\textbf{Quantitative evaluation of customized multi-concept generation methods on MC++.} The metrics are first computed for each composition then averaged. CLIP-T measures prompt fidelity. CLIP-I and DINO measure subject fidelity.
	}
	\label{tab:multiconceptresult}
\end{table*}
	
\begin{table}
	\scriptsize
	\centering
	\begin{tabular}{cccc}
		\toprule
		\textbf{Method} & \textbf{Full Prompt\textuparrow} & \makebox[0.06\textwidth]{\textbf{Min. Object\textuparrow}} & \textbf{Text-Text\textuparrow} \\
		\midrule
		Composable Diffusion \cite{liu2022compositional}     & 0.327 & 0.250 & 0.738 \\
		Attend-and-Excite \cite{Chefer2023AttendandExciteAS} & 0.352 & 0.264 & 0.818 \\
		Divide-and-Bind \cite{{DBLP:conf/bmvc/LiK0K23}}      & 0.347 & 0.258 & 0.817 \\
		\rowcolor{gray!15}
		Ours                                                 & \textbf{0.356} & \textbf{0.266} & \textbf{0.833} \\
		\bottomrule
	\end{tabular}
	\caption{\textbf{Quantitative evaluation of compositional generation methods.} Metrics include \textit{Full Prompt Similarity},\textit{ Minimum Object Similarity} and \textit{Text-Text Similarity}. 
	}
	\label{tab:gsnresult}
\end{table}
		
\subsection{Customized Multi-concept Generation Results}
		
\noindent
\textbf{Qualitative evaluation.}

\Cref{fig:comparison} shows the generated images of our method and the baselines, containing compositions of two to four concepts. In the first row, we show the unreal composition of a cat wearing a jacket. Ours demonstrates a better consistency with both the cat and the jacket.
In the fourth row, ours demonstrates superior editability compared to the baselines, successfully edits the chair to purple while not affecting the shape and color of the table.
In the second to last row, we edit the artistic style of the image.
In the last row, we show results of composing four concepts. The baselines failed to depict all the four concepts.
The joint training of Custom Diffusion and model merging of Mix-of-Show sometimes make the concepts merge. As shown in the third and fourth row, the concepts would share some features. The Cones~2 and Textual Inversion have inferior concept consistency due to their architecture design that only an embedding is trained for each concept. For better consistency, more tunable parameters are needed. Ours avoids feature leakage between concepts in the training stage and the MCG provides a new way to compose them in the inference stage while reducing feature leakage.

\noindent
\textbf{Quantitative evaluation.}
We generate 16 samples per prompt with the same set of random seeds, resulting in a total of 33,024 images for each method. As shown in \cref{tab:multiconceptresult}, our method outperforms the baselines in terms of prompt fidelity and subject fidelity. In the three and two concepts setting, Cones~2 demonstrates a higher prompt fidelity, which may come from its residual embedding design. The design reduces the impact of the customized concepts on the base model. However, a relatively small parameter size also hinders the learning of new concepts. Ours strikes a better balance between prompt fidelity and subject fidelity overall.

\subsection{Compositional Generation Results}

\noindent
\textbf{Qualitative evaluation.}
\Cref{fig:comparison-gsn} shows the generated images of our method and the baselines.
Composable Diffusion~\cite{liu2022compositional} sometimes merges the two objects, \eg the bench made of blue ballon.
Attend-and-Excite~\cite{Chefer2023AttendandExciteAS} and Divide-and-Bind~\cite{{DBLP:conf/bmvc/LiK0K23}} come across incorrect attribute binding problem, \eg the rabbit with ivories, the blue bench.
While our method successfully generates the objects without confusing the attributes belonging to each object.

\noindent
\textbf{Quantitative evaluation.}
We generate 16 samples per prompt with the same set of random seeds, resulting in a total of 4,416 images for each method. As shown in \cref{tab:gsnresult}, our method outperforms the baselines.

\subsection{User Study}

\begin{table}
	\scriptsize
	\centering
	\begin{tabular}{ccc|cc}
		\toprule
		{\textbf{Method}} & \makebox[0.03\textwidth]{\textbf{Text Align.}} & \makebox[0.07\textwidth]{\textbf{Image Align.}} & \textbf{Method} & \makebox[0.03\textwidth]{\textbf{Text Align.}} \\
		\midrule
		Custom Diff. & 3.66\% & 7.77\% &             Composable Diff. & 1.19\% \\
		Mix-of-Show      & 8.66\% & 16.51\% &            Attend-and-Excite   & 10.00\% \\
		Cones 2          & 8.93\% & 2.32\% &             Divide-and-Bind     & 6.55\% \\
		\rowcolor{gray!15}
		Ours & \textbf{78.75\%} & \textbf{73.39\%} &     Ours                & \textbf{82.26\%} \\
		\bottomrule
	\end{tabular}
	\caption{\textbf{User study.} Twenty-eight users take part in the survey. The users are asked to select the image most consistent with the prompt (Text Align.) and the reference images (Image Align.).
	}
	\label{tab:userstudy}
\end{table}
We perform a user study to evaluate our method. For customized multi-concept generation, we show users the generated images of each method along with two reference images of the concept. Then the users are asked to select the image that is most consistent with the prompt and the image that best represents the concepts. For compositional generation, we ask them to choose the one most consistent with the prompt. The questionnaire contains 40 questions for the customization task and 30 questions for compositional generation. We collect 28 responses in total. As shown in \cref{tab:userstudy}, our method outperforms the baselines.

\subsection{Ablation Study}
\begin{figure}
	\scriptsize
	\centering
	\includegraphics[width=\linewidth]{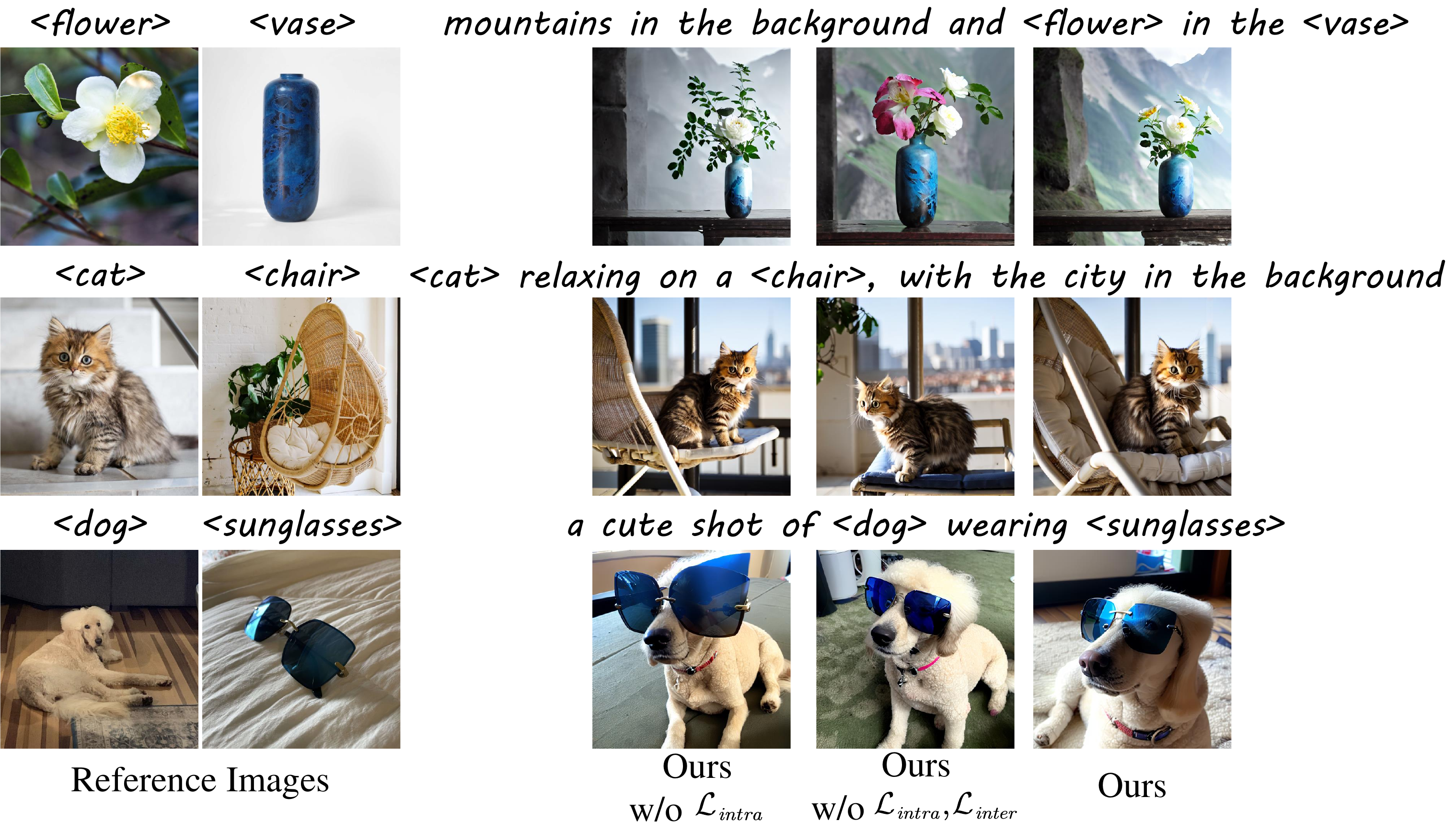}
	\caption{\textbf{Qualitative comparisons of ablation variants.} Omitting any of the components harms the generation quality.
	}
	\label{fig:ablation}
\end{figure}

The ablation for customized multi-concept generation is carried out on a subset of MC++. For compositional generation, we use the full benchmark dataset mentioned in \cref{sec:dataset}.
We show ablation of our proposed MCG and attention grounding training in \cref{tab:ablation}, \cref{fig:ablation}. 

\begin{table}
	\scriptsize
	\centering
	\begin{tabular}{lccc|ccc}
		\toprule
		\multirow{2}{*}{{\textbf{Method}}} & \multicolumn{3}{c|}{\textbf{Customization}} & \multicolumn{3}{c}{\textbf{Compositional Gen.}} \\
		~ & \makebox[0.03\textwidth]{CLIP-T\textuparrow} & \makebox[0.01\textwidth]{CLIP-I\textuparrow} & \makebox[0.01\textwidth]{DINO\textuparrow} & \makebox[0.01\textwidth]{Full.\textuparrow} & \makebox[0.01\textwidth]{Min.\textuparrow} & \makebox[0.01\textwidth]{T-T\textuparrow} \\
		\midrule
		\rowcolor{gray!15}
		Ours & \textbf{0.789} & \textbf{0.716} & \textbf{0.434} & \textbf{0.356} & \textbf{0.266} & \textbf{0.833}\\
		$- \text{Attn grounding}$ & 0.784 & 0.712 & 0.426 & $-$ & $-$ & $-$ \\
		$-\mathcal{L}_{intra}$ & 0.779 & 0.706 & 0.418 & 0.355 & 0.264 & 0.830 \\
		$-\mathcal{L}_{inter}$ & 0.440 & 0.539 & 0.356s & 0.353 & 0.263 & 0.827 \\
		\bottomrule
	\end{tabular}
	\caption{\textbf{Ablation Study.} Ablation of the two loss terms in MCG and attention grounding training.
	}
	\label{tab:ablation}
\end{table}

\subsection{Limitations}

\begin{figure}[tb]
	\centering
	\includegraphics[width=\linewidth]{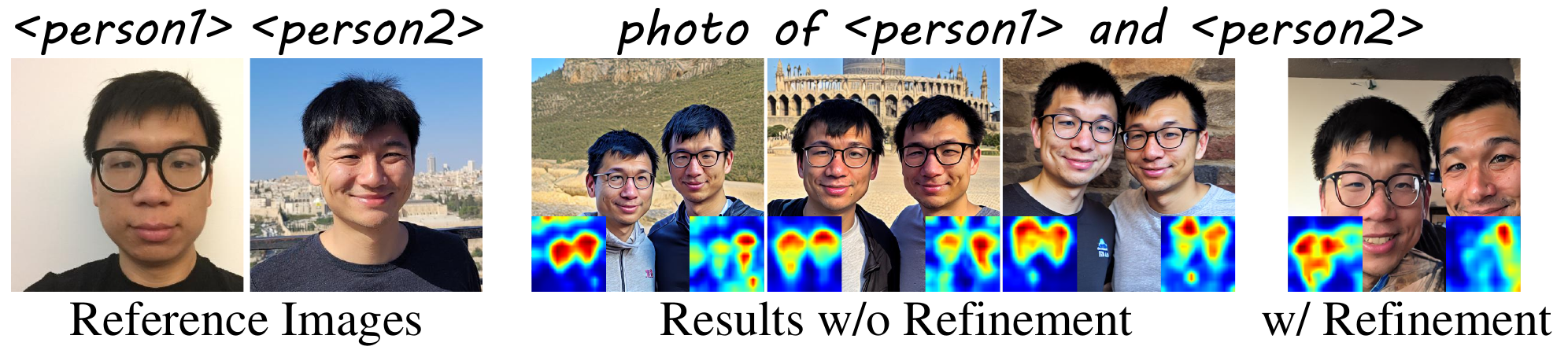}	
	\caption{\textbf{Limitations.}
		Failure cases on composing two people.
	}
	\vspace{-12pt}
	\label{fig:limit}
\end{figure}

In \cref{fig:limit}, our method fails to compose two people, \eg mixing their identities such as glasses. The attention maps of selected tokens are not sufficient for distinguishing the identities. With a refinement stage, the two people can be distinguished. Basically we give different weights to the regions of different concepts when merging the outputs of the unets. We do not include the refinement stage for the simplicity of our method. Details of the refinement stage are in \textit{suppl}.

\section{Conclusion}
We have proposed MC$^2$ which allows combination of various heterogeneous single-concept customized models, enabling a more flexible customized multi-concept generation with improved fidelity. Multiple customized models are seamlessly integrated with Multi-concept Guidance to synthesize a natural composition of the multiple customized concepts. MC$^2$ can further be extended to elevate the compositional capabilities of text-to-image generation.
{
    \small
    \bibliographystyle{ieeenat_fullname}
    \bibliography{main}
}

\clearpage
\setcounter{page}{1}
\maketitlesupplementary

In \cref{sec:benchmark}, we show the details of the MC++ benchmark. In \cref{sec:details}, we show more implementation details, including hyperparameter setting and corresponding ablation. In \cref{sec:limitations}, we discuss the refinement stage mentioned in Sec.~\textcolor{cvprblue}{4.7} of the main paper and more possible limitations. In \cref{sec:results}, we show additional results. In \cref{sec:future} and \cref{sec:impact}, we discuss future work and societal impact of our work.

\section{MC++ Benchmark}
\label{sec:benchmark}

We adapt CustomConcept101~\cite{kumari2023multi} to compositions of three and four subjects. The original CustomConcept101 is a dataset of 101 concpets with 3-15 images for each concept. For evaluating multi-concept customization, the CustomConept101 contains prompts for 101 compositons of two subjects. To facilitate the evaluation of multi-concept customization on more than two subjects, we collect prompts for 57 compositions of three subjects, and 14 compositions of four subjects from the original CustomConcept101. Each composition has 12 prompts. Basicly we follow the procedure of \cite{kumari2023multi} to get the prompts. We first used ChatGPT~\cite{chatgpt} to propose the prompts then manually filter them to get the final set of prompts.

\section{Implementation Details}
\label{sec:details}

We choose LoRA as our single-concept customization model. We adopt a popular community implementation of LoRA~\cite{sdscripts}. LoRA modules are trained for linear layers and $1 \times 1$ conv layers in text encoder and unet with rank set to 16. We do not finetune the token embeddings. All reference images are captioned as \texttt{photo of a <concept name>}. Each concept has a unique \texttt{<concept name>} defined by CustomConcept101, \eg \texttt{pet\_dog1}. All LoRAs are trained for 1000 steps with batch size set to 2, learning rate set to 1e-4. LoRA scale is set to 0.7 for merging the LoRA parameters into the diffusion model during inference. 

\begin{figure}
	\centering
	\includegraphics[width=\linewidth]{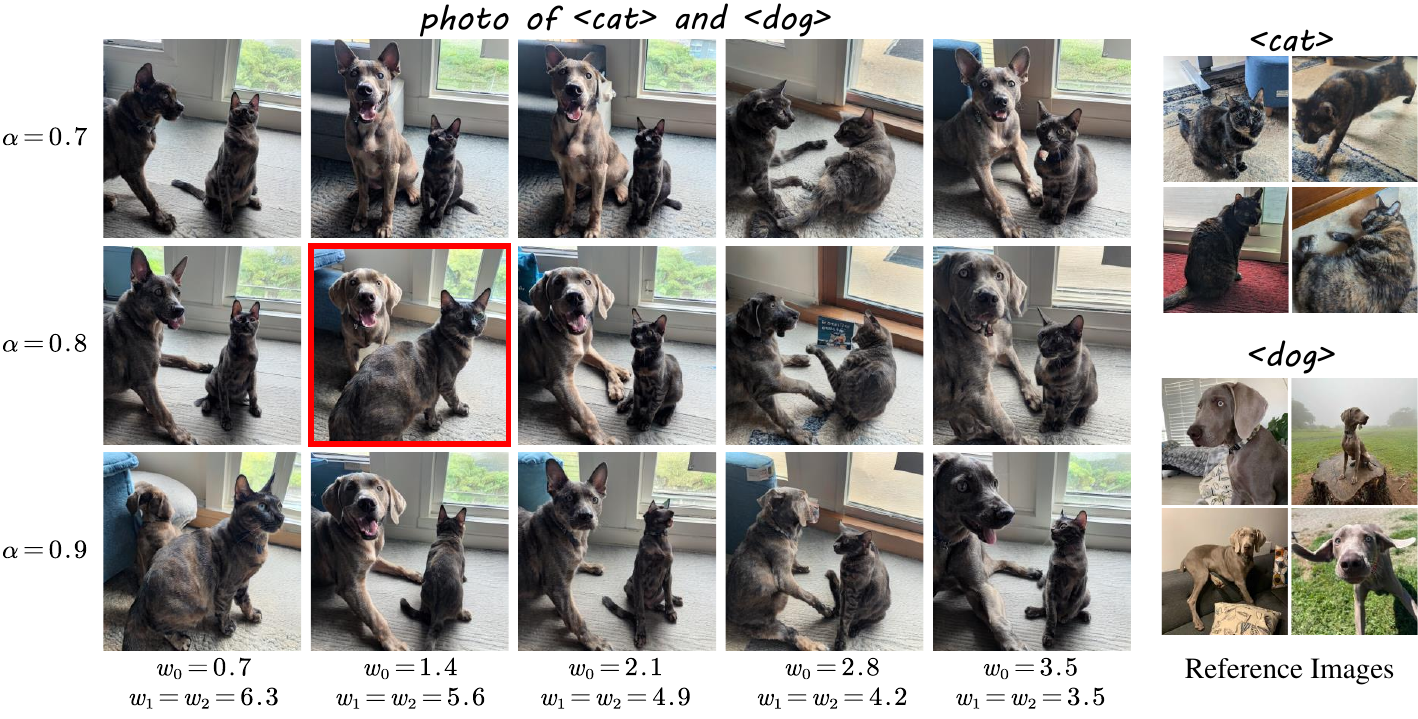}
	\caption{
		\textbf{Abltion of hyperparameters of our method.} Ablation of hyperparameters $\alpha, w_0, w_1, w_2$. 
	}
	\label{fig:hyperpara-ablation}
\end{figure}

Here we list the hyperparameters used in our method.
In Eq.~(\textcolor{cvprblue}{7}), $\alpha$ is set to 0.8. In Eq.~(\textcolor{cvprblue}{8}), we schedule the learning rate $\lambda$ linearly. It starts from 20 then decays to 10 linearly. We take a single gradient step per diffusion time step. In Eq.~(\textcolor{cvprblue}{9}), $w_0, w_1,... , w_n$ are set to $1.4, 5.6,... , 5.6$ for customized multi-concept generation, set to $5.6, 1.4,... , 1.4$ for compositional generation. In Eq.~(\textcolor{cvprblue}{10}), $\alpha_1, \alpha_2$ are set to 0.5, 0.4. The MCG is performed at the first 25 steps of the diffusion process. The Gaussian filter used to smooth the cross-attention maps has a kernel size of 3 and a standard deviation of 0.5. \Cref{fig:hyperpara-ablation} shows ablation of the hyperparameters. The $\alpha$ controls the balance between $\mathcal{L}_{inter}$ and $\mathcal{L}_{intra}$. The $w$ controls the balance between the uncustomized model and the customized models. The \textcolor{red}{red box} marks our recommended hyperparameters.

\begin{figure}
	\centering
	\includegraphics[width=\linewidth]{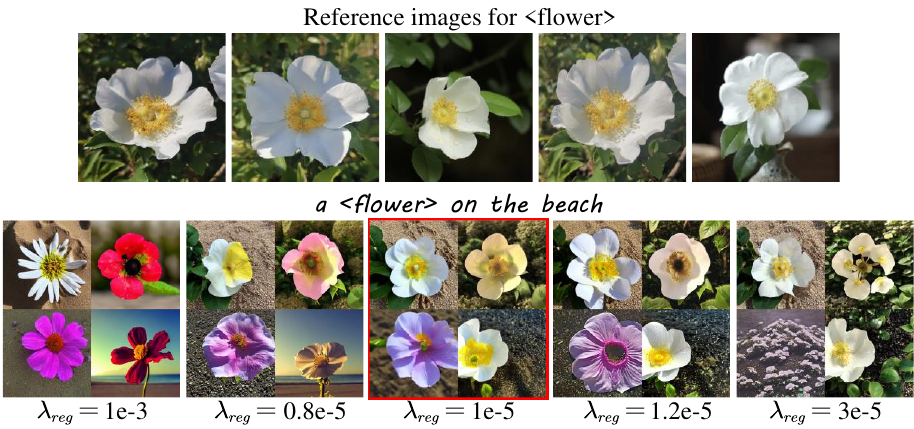}
	\caption{
		\textbf{Ablation of the parameter $\lambda_{reg}$ in Cones~2~\cite{liu2023cones}.} The $\lambda_{reg}$ is the coefficient of the loss term $\mathcal{L}_{reg}$ for training a Cones~2 model.
	}
	\label{fig:cones2-ablation}
\end{figure}

About the implementation of the baselines, we basically follow their official repo except Cones~2~\cite{liu2023cones}. It was implemented based on SD~2.1 rather than SD~1.5. As our experiments are based on SD~1.5, we adapt the official code to SD~1.5. However, we found their default hyperparameter sub-optimal for SD~1.5. As shown in \cref{fig:cones2-ablation}, when the $\lambda_{reg}$ is set to default value 1e-3, the model failed to learn the \texttt{<flower>} concept. The $\lambda_{reg}$ is the coefficient of the regularization term $\mathcal{L}_{reg}$ for training a Cones~2 model. A large $\lambda_{reg}$ hinders the learning of the model. We set $\lambda_{reg}$ to 1e-5 in the experiments.

\section{Limitations}
\label{sec:limitations}

\begin{algorithm}
	\caption{Propose Masks Based on Cross-Attn Maps}
	\label{alg:mask}
	\begin{algorithmic}[1]
		\Statex \hspace*{-\leftmargin}\textbf{Input:} Cross-attention maps $A_1, A_2$ for the two subjects, overlap threshold $\theta_1$, binarization threshold $\theta_2$.
		\Statex \hspace*{-\leftmargin}\textbf{Output:} Masks $M_1, M_2$ for the two subjects
		\State $A_1 \gets \text{Gaussian}((A_1 - \min(A_1)) / \max(A_1))$ \label{line:rescale1}
		\State $A_2 \gets \text{Gaussian}((A_2 - \min(A_2)) / \max(A_2))$ \label{line:rescale2}
		\State $M_1, M_2 \gets A_1 > A_2, A_2 > A_1$ \label{line:overlap1}
		\State $O \gets (A_1 > \theta_1) \& (A_2 > \theta_1)$ \label{line:overlap2}
		\State $M_1 \gets M_1 \odot (1 - O\& M_2)$ \label{line:overlap3}
		\State $M_2 \gets M_2 \odot (1 - O\& M_1)$ \label{line:overlap4}
		\State $M_1, M_2 \gets M_1 > \theta_2, M_2 > \theta_2$ \label{line:binarize}
		\State $M \gets \text{Dilate}(M_1 | M_2)$ \label{line:dilate}
		\State $M \gets M - M_1 | M_2$
		\State $D_1 \gets \text{DistanceTransform}(1 - M_1)$
		\State $D_2 \gets \text{DistanceTransform}(1 - M_2)$
		\State $M_1 \gets M \odot (D_1 < D_2) + M_1$
		\State $M_2 \gets M \odot (D_2 < D_1) + M_2$ \label{line:dilate2}
		\State $M_1 \gets \text{Gaussian}(M_1)$
		\State $M_2 \gets \text{Gaussian}(M_2)$
		\State \Return $M_1, M_2$
	\end{algorithmic}
\end{algorithm}

\begin{figure}
	\centering
	\includegraphics[width=\linewidth]{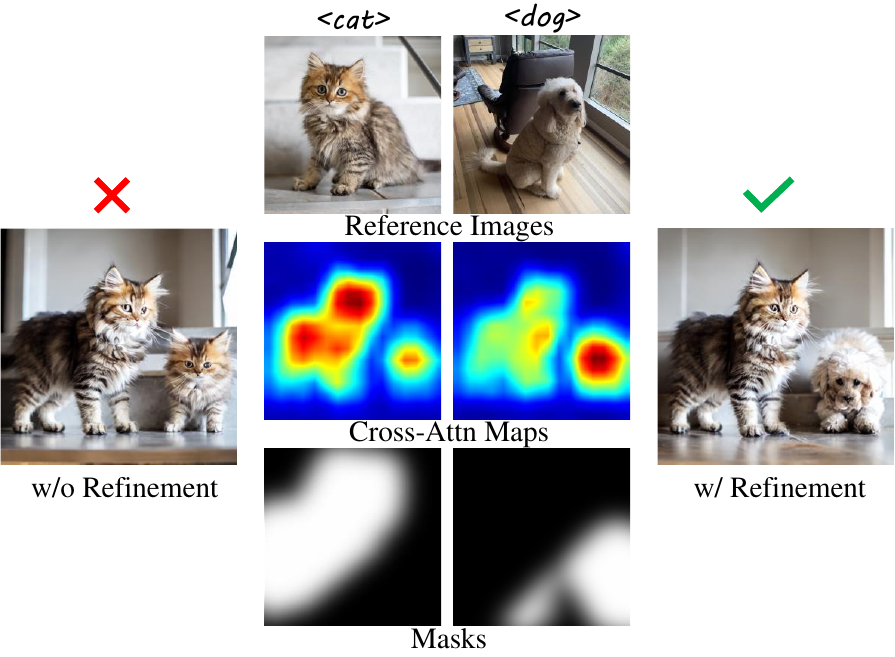}
	\caption{
		\textbf{Visualization of the refinement stage.} Here we show the cross-attention maps and masks involved in the refinement stage.
	}
	\label{fig:mask}
\end{figure}

As mentioned in Sec.~\textcolor{cvprblue}{4.7} of the main paper, we add a refinement stage for the composition of multiple concepts that share similar features. When two concepts share similar features, it may be difficult to tell them apart based on the corresponding cross-attention maps. Here we propose an algorithm for extracting masks for the concepts based on the cross-attention maps. The goal is to propose a binary mask with soft boundary for each concept. \Cref{alg:mask} shows how to extract masks for two subjects. We visualize the masks in \cref{fig:mask}. In \cref{line:rescale1,line:rescale2}, we rescale and smooth the attention maps. In \cref{line:overlap1,line:overlap2,line:overlap3,line:overlap4}, we detect the overlap of the two attention maps with a threshold. In \cref{line:binarize}, we binarize the attention maps to get the masks. From \cref{line:dilate} to \cref{line:dilate2}, we dilate the masks and assign the dilated area to the proper subject based on distance transform. Finally we perform Gaussian blur to get a soft boundary.
The masks $M_i$ are then used in \cref{eq:merging} for merging the outputs of multiple diffusion models, which is identical to \cite{BarTal2023MultiDiffusionFD}. 
\begin{equation}
	z_{t-1} = z_{t-1}^{u} + \displaystyle\sum_{i=0}^{n}w_i(M_i \odot z_{t-1}^{i}-z_{t-1}^{u}).
	\label{eq:merging}
\end{equation}
Note that the masks are generated on the fly and we replace the Eq.~(\textcolor{cvprblue}{9}) in the main paper with \cref{eq:merging}. As shown in the \cref{fig:mask}, with the refinement stage, the dog is correctly generated.

Another limitation of our method is that the architecture of multiple parallel diffusion models requires relatively larger memory usage, particularly when composing multiple customized concepts. This is partly due to our implementation. We literally maintain multiple instances of the same diffusion model in memory for the sake of simplicity. Addressing this limitation involves optimizing memory utilization by storing only a single instance of the diffusion model in memory, thereby enhancing memory efficiency.

Despite MC$^2$ enables users to compose multiple separately trained, even heterogeneous customized models, the customized models should be trained from the same diffusion model. Such limitation is inherited from \cite{liu2022compositional}.

\section{Additional Results}
\label{sec:results}

\begin{figure}
	\centering
	\begin{overpic}[width=\linewidth]{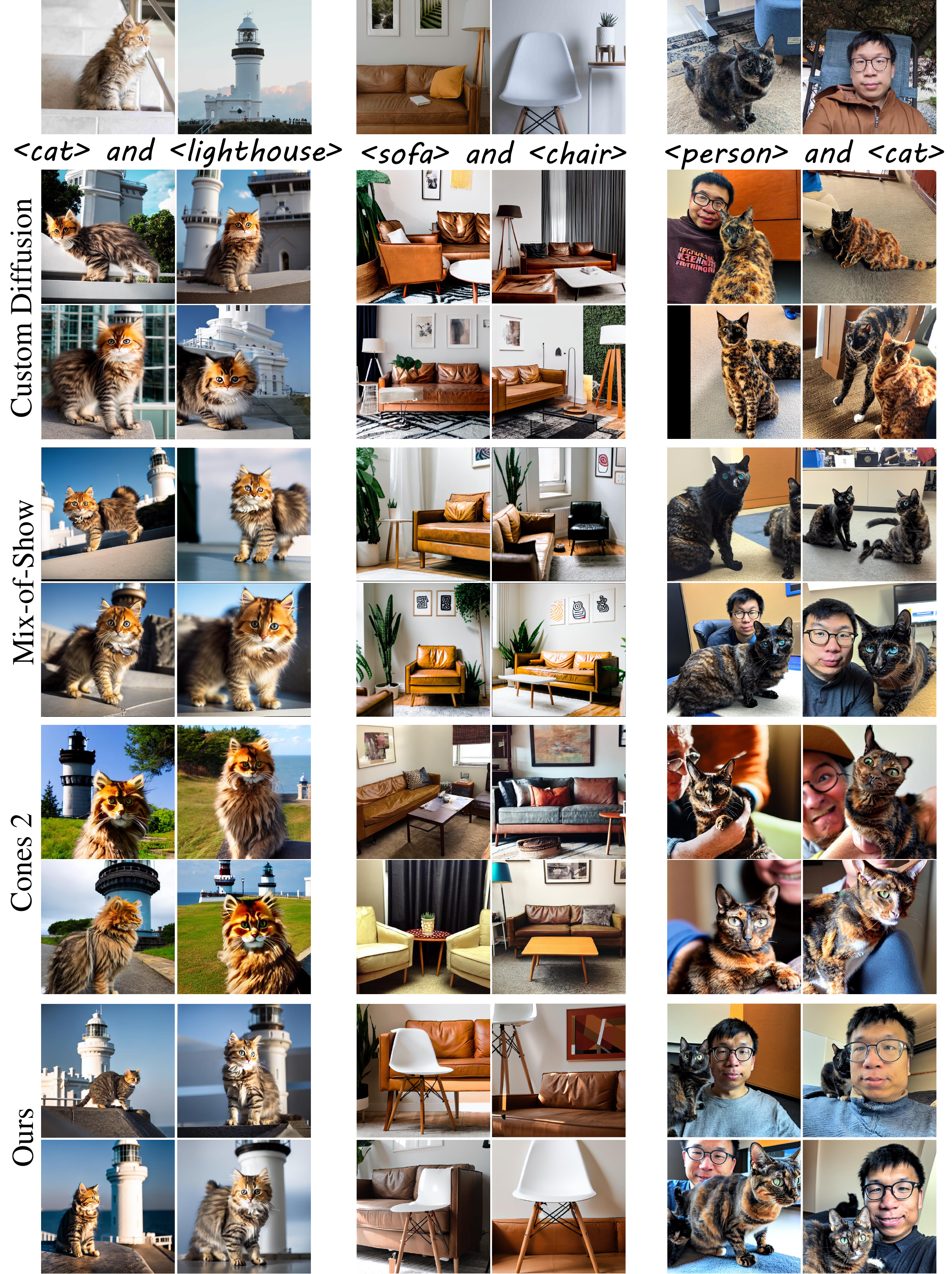}
		\scriptsize
		\begin{sideways}
			\put(84.5, -2.5){\cite{kumari2023multi}}
			\put(61, -2.5){\cite{gu2023mixofshow}}
			\put(36.2, -2.5){\cite{liu2023cones}}
		\end{sideways}
	\end{overpic}
	\caption{\textbf{Comparisons on generating two subjects.} Qualitative comparisons of customized multi-concept generation methods.
	}
	\label{fig:comparison-custom}
\end{figure}

\begin{figure}
	\centering
	\begin{overpic}[width=0.85\linewidth]{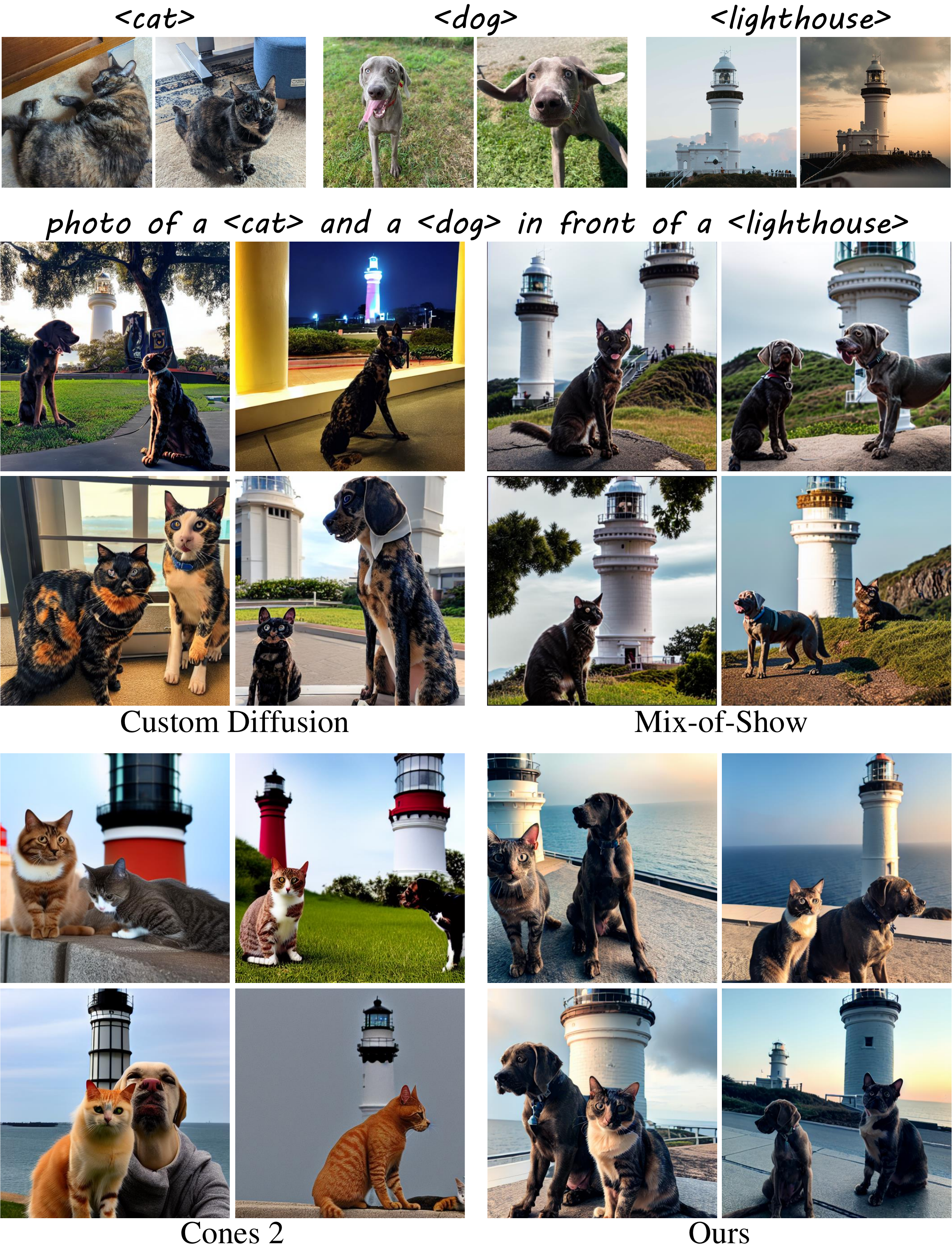}
		\scriptsize
		\put(28, 41.6){\cite{kumari2023multi}}
		\put(64.5, 41.6){\cite{gu2023mixofshow}}
		\put(22.7, 0.9){\cite{liu2023cones}}
	\end{overpic}
	\caption{
		\textbf{Comparisons on generating three subjects.} Qualitative comparisons of customized multi-concept generation methods on three subjects.
	}
	\label{fig:comparison-custom-three}
\end{figure}

\Cref{fig:comparison-custom} shows more qualitative comparisons of the proposed method and the baselines \cite{liu2023cones,kumari2023multi, gu2023mixofshow} on customized multi-concept generation.
The baselines sometimes omit one of the specified concepts, \eg the white chair for \texttt{<sofa> and <chair>} and the person for \texttt{<person> and <cat>}. Our method demonstrates higher fidelity to the reference images even compared to Custom Diffusion \cite{kumari2023multi} that requires jointly training the two concepts, or Mix-of-Show~\cite{gu2023mixofshow} that requires training to merge the two single-concept customized models. Cones~2 \cite{liu2023cones} shows relatively low fidelity to the reference images, considering that it requires the least trained parameters.
Our method demonstrates a more satisfying effect.
\Cref{fig:comparison-custom-three} shows qualitative comparisons of the customized multi-concept generation methods on composition of three concepts. Our method demonstrates more higher fidelity to the reference images.

\begin{figure}
	\centering
	\begin{overpic}[width=0.85\linewidth]{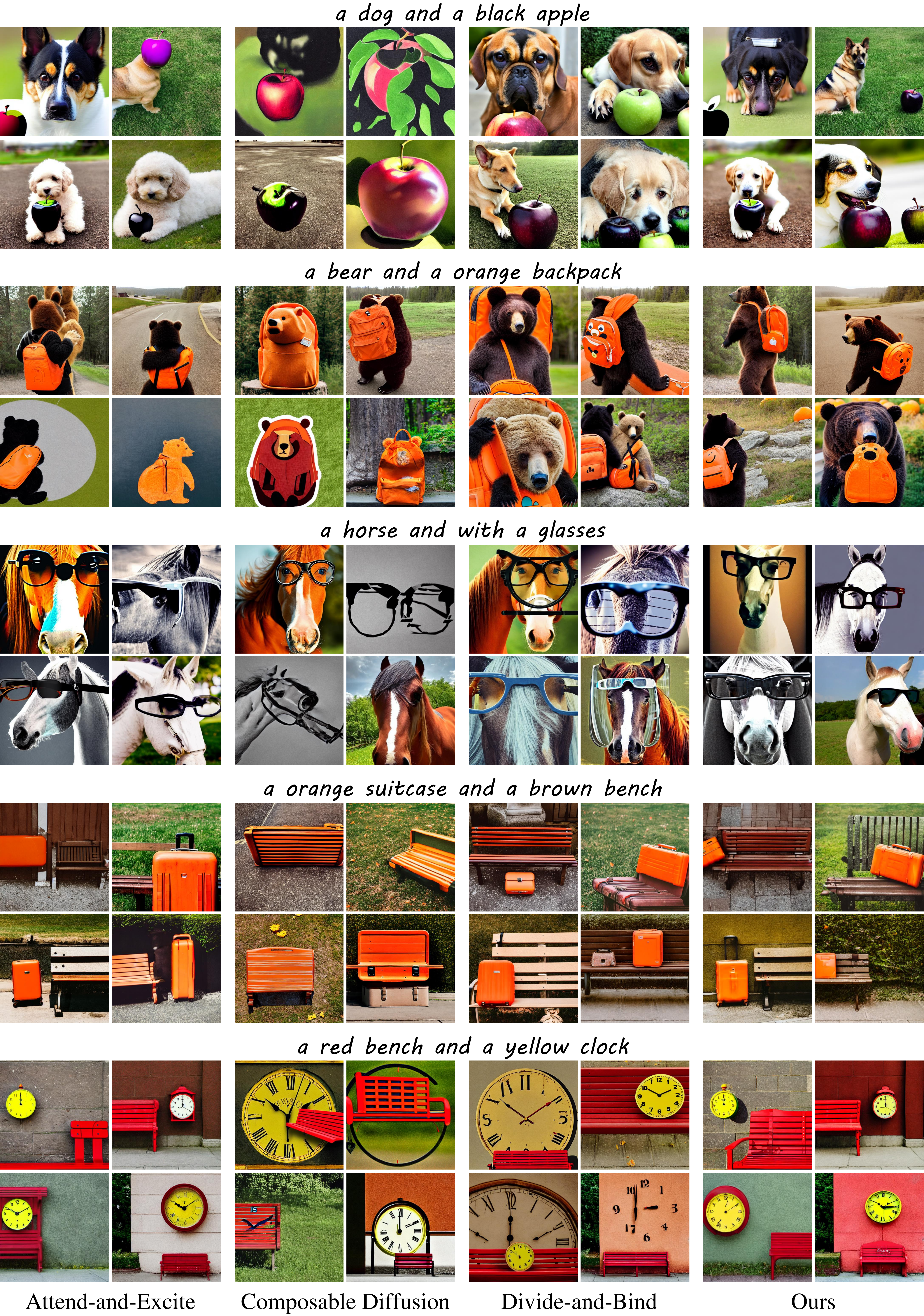}
		\scriptsize
		\put(15, 0.5){\cite{Chefer2023AttendandExciteAS}}
		\put(34.3, 0.5){\cite{liu2022compositional}}
		\put(50, 0.5){\cite{DBLP:conf/bmvc/LiK0K23}}
	\end{overpic}
	\caption{
		Qualitative comparisons of compositional generation methods.
	}
	\label{fig:comparison-gsn-supply}
\end{figure}

\Cref{fig:comparison-gsn-supply} shows more qualitative comparisons of the compositional generation methods.
Our method demonstrates better consistency with the input text prompts compared to the baselines \cite{Chefer2023AttendandExciteAS, liu2022compositional, DBLP:conf/bmvc/LiK0K23}.

\section{Future Work}
\label{sec:future}

In addition to addressing the limitations mentioned in \Cref{sec:limitations}, there exist several promising avenues for future research.
\cite{chen2023videodreamer, wang2024customvideo} delve into the realm of multi-concept customization for text-to-video generation. An intriguing prospect is to investigate the adaptability of our proposed MC$^2$ to the domain of text-to-video generation.

For compositional generation, an interesting avenue for exploration involves building upon our methodology. Our approach not only addresses current challenges but also opens up a novel design space for further investigation. This provides a foundation for the development of innovative methods to enhance compositional generation techniques.

\section{Societal Impact}
\label{sec:impact}

First and foremost, MC$^2$ empowers users to effortlessly generate visually captivating compositions reflecting their unique ideas and preference.
Additionally, MC$^2$'s ability to enhance the capabilities of existing text-to-image diffusion models opens up new avenues for artistic exploration and innovation, potentially inspiring broader adoption and engagement in creative endeavors.
However, MC$^2$ may blur the lines between ethical and unethical image manipulation. Without proper guidance and ethical considerations, individuals may engage in harmful practices such as image forgery or digital impersonation. We advocate for the development of legal frameworks that address AI-generated content, including penalties for malicious use.

\end{document}